%% file: root.tex
\documentclass[letterpaper, 10 pt, conference]{ieeeconf}  

\IEEEoverridecommandlockouts                              

\usepackage{graphicx}
\usepackage{amsmath}
\usepackage{amssymb}
\usepackage{booktabs}
\usepackage{multirow}
\usepackage{cite}
\usepackage{algorithm}
\usepackage{algorithmic}
\usepackage[hidelinks]{hyperref}

\title{\LARGE \bf
EXE: Egocentric Cross-Embodiment Pretraining \\
via Factorized Latent Actions for VLA Models 
}

\author{Qingyuan Cai$^{1,2,\dagger}$, Bo Zhang$^{1}$, Yongzhen Huang$^{2}$, and Jianyu Chen$^{1,3}$%
\thanks{$^{1}$ROBOTERA.}%
\thanks{$^{2}$Beijing Normal University.}%
\thanks{$^{3}$Tsinghua University.}%
\thanks{$\dagger$This work was done during Qingyuan Cai’s internship at ROBOTERA.}%
}

\begin{document}

\maketitle
\thispagestyle{empty}
\pagestyle{empty}

\begin{abstract} 
Vision-Language-Action (VLA) models are bottlenecked by the scarcity of robot data. Egocentric human manipulation videos offer an abundant, diverse, and semantically-rich alternative that inherently captures manipulation intent. However, the embodiment gap makes it difficult to extract action-relevant representations from these videos. From the perspective of action execution, human hands and robots share analogous manipulation primitives---reaching, grasping, placing---yet exhibit distinct kinematics tied to their embodiments. This might motivate our key approach: egocentric video pretraining can yield shared action representations capturing transferable manipulation skills, while human-specific motion patterns should be disentangled to avoid interfering with robot policy learning. To this end, we propose EXE, an \emph{E}gocentric Cross(\emph{X})-\emph{E}mbodiment pretraining framework that learns a \emph{factorized latent action} representation from egocentric human videos, separating shared manipulation priors from embodiment-specific signals. By leveraging the shared action space, the robot policy acquires generalizable manipulation semantics from diverse human demonstrations without being biased by human-specific kinematics. This shared knowledge transfers across embodiments with different kinematic structures, enabling the same pretraining to benefit both parallel grippers and dexterous hands. EXE yields consistent gains over baseline without pretraining on RoboTwin 2.0, RoboCasa GR1, and real-world manipulation tasks.
\end{abstract}

\section{Introduction}

Vision-Language-Action (VLA) models unify visual perception, language understanding, and action generation in a single architecture, enabling robots to follow open-ended instructions across diverse manipulation tasks~\cite{brohan2023rt2,kim2024openvla,black2025pi0,octo2024}. Their success, however, is tightly coupled to the scale of robot demonstration data. Curating large manipulation datasets requires physical teleoperation or scripted policies on real hardware, a process that is expensive, slow, and inherently limited in scene and task diversity~\cite{openxembodiment2024}. This data bottleneck restricts VLA generalization and has motivated a search for alternative, scalable supervision sources.

Egocentric human videos offer a compelling alternative. Datasets such as Ego4D~\cite{grauman2022ego4d}, Epic-Kitchens~\cite{damen2022epickitchens}, EgoDex~\cite{hoque2026egodex}, VITRA-1M~\cite{li2025scalable}, and EgoLive~\cite{li2026egolive,zhang2026joyai} capture thousands of hours of first-person hand manipulation with rich task semantics, diverse objects, and natural language annotations. Because these videos share the egocentric perspective of a robot's wrist or head camera, they natively encode manipulation intent. Several works exploit this connection: GR-1~\cite{wu2024gr1} and GR-2~\cite{cheang2025gr2} pretrain video-generation backbones on human video before fine-tuning for robot control, and HPT~\cite{wang2024hpt} maps heterogeneous embodiment streams through a shared transformer trunk. Yet these approaches consume visual features only and do not attempt to transfer the \emph{action} information embedded in human hand motions. The open question is how to convert action-level knowledge in egocentric video into a form that a VLA can absorb during pretraining and later translate into executable robot actions.

From the perspective of action execution, humans and robots share analogous manipulation primitives such as reaching, grasping, and placing, yet realize them with distinct kinematics---human hands parameterized by MANO finger joints~\cite{romero2017mano} versus robot end-effector poses with gripper or dexterous-hand configurations---making direct alignment across the two action spaces difficult. Two strategies have been tried to bridge this gap, each with a fundamental limitation. The first, exemplified by UniT~\cite{chen2026unit}, forces the entire action into a single shared codebook, so the shared codes inevitably absorb embodiment-specific geometry that obscures the transferable task semantics. The second, exemplified by LAPA~\cite{ye2025lapa}, learns latent actions from visual frame transitions, but raw observations entangle the action signal with task-irrelevant content such as background and lighting---leading CLAP~\cite{zhang2026clap} to note that such latent actions encode visual noise rather than manipulation skill.

These limitations reveal two questions at the heart of cross-embodiment transfer: \emph{which} parts of an action are worth sharing, and \emph{how} shared actions should be aligned across embodiments. Our work starts from two observations that address them respectively. \textbf{Observation~1:} \emph{Certain dimensions of the action transfer better across embodiments, while others are inherently difficult to transfer.} Specifically, the 3D position of the wrist (for humans) or end-effector (for robots) encodes spatial task intent---reaching, trajectory following, target approaching, which is shared across embodiments regardless of kinematic structure. Finger joint configurations and gripper geometry, by contrast, are embodiment-specific and do not generalize across domains. This motivates our feature-group routing design: rather than unifying the entire action, which inevitably introduces noise, we factorize it into a shareable positional subspace that captures cross-embodiment task semantics and an embodiment-specific kinematic subspace kept private to each domain.

\textbf{Observation~2:} Given a shared positional subspace, the next question is how to align representations across embodiments within it. \emph{Actions should not only be aligned when their spatial trajectories are similar, but also when they carry the same task semantics despite dissimilar trajectories.} A contrastive objective such as InfoNCE~\cite{oord2018representation} naturally handles the first case: it pulls together cross-domain samples with similar spatial trajectories and pushes apart those with dissimilar ones. However, spatial similarity does not imply semantic equivalence---actions that share the same task semantics may still follow substantially different trajectories, and InfoNCE would wrongly drive such pairs apart. For example, a human grasping a cup from above with five fingers and a robot clamping the same cup from the side with a parallel gripper both execute ``grasp the cup,'' yet their spatial trajectories differ substantially. These are exactly the pairs a cross-embodiment representation should unite. This motivates our \emph{language-guided relational distillation}: language provides an external semantic signal invariant to embodiment geometry, supplying pairwise similarity as a soft target to align semantically equivalent but spatially dissimilar actions in the shared space, complementing the contrastive spatial alignment.


We present \textbf{EXE}, \textbf{E}gocentric Cross(\textbf{X})-\textbf{E}mbodiment Pretraining via Factorized Latent Actions, a framework that turns these observations into a practical VLA pretraining pipeline. EXE introduces three components:
\begin{itemize}
    \item A \textbf{feature group-routed action tokenizer} that factorizes each action into a shareable positional subspace (3D wrist/end-effector position) and an embodiment-specific kinematic subspace (finger joints or gripper). The shareable subspace is quantized through a shared codebook trained jointly on human and robot data, while each domain maintains its own codebook for embodiment-specific features. This isolates transferable task semantics from domain-specific geometry within a single quantization framework.
    \item A \textbf{dual cross-domain alignment} scheme for the shared codebook that combines contrastive learning with language-guided relational distillation. Contrastive learning pulls together cross-domain actions with similar spatial trajectories, while frozen language embeddings~\cite{zhang2025qwen3} supply pairwise semantic similarity as a soft target to further align semantically equivalent yet spatially dissimilar actions, which contrastive learning alone would drive apart.
    \item A \textbf{three-stage pipeline}: (1)~training the factorized tokenizer on combined human and robot action data to establish shared and private codebooks; (2)~training the VLM models with variable egocentric human video and robot data, supervising the action head with discrete tokens from the shared vocabulary so that the model absorbs cross-embodiment manipulation knowledge; (3)~fine-tuning on downstream robot tasks, where pretrained knowledge transfers directly through the shared action codes.
\end{itemize}

Since the shared knowledge can be constructed across embodiments with different kinematic structures, EXE benefits both gripper (RoboTwin 2.0~\cite{chen2025robotwin2}) and dexterous-hand (RoboCasa GR1~\cite{nvidia2025gr00tn1openfoundation}), as well as real-world tasks, consistently improving success rates over baselines without pretraining. Ablation studies confirm that feature-group factorization and language-guided alignment each contribute independently.

\section{Related Work}
\paragraph{Egocentric human video}
Egocentric video captures manipulation from the operator's own viewpoint, providing first-person hand--object interactions at a scale and diversity unreachable by robot teleoperation. Large-scale datasets~\cite{grauman2022ego4d,grauman2024ego,damen2022epickitchens,goyal2017something,buildaiegocentric10k2025,hoque2026egodex, cai2026barbiegait} accumulate thousands of hours of daily activities with dense annotations spanning ego/exo views, action labels, and hand-pose tracking, driving progress in action recognition, hand--object interaction, and affordance learning. Notably, the egocentric viewpoint aligns naturally with a robot's wrist or head camera and inherently encodes manipulation intent, making it an increasingly attractive supervision source for robot learning.

\paragraph{Egocentric video for VLA pretraining}
A growing line of work pretrains VLA models on egocentric human video before robot fine-tuning. The key distinction among methods lies in \emph{how human motion is converted into a robot-executable action signal}. \textbf{Explicit} approaches recover 3D poses and map them to robot actions via kinematic retargeting or inverse kinematics~\cite{yang2025egovla,mu2026deximit,zheng2026egoscale,zhang2026unidex,zhong2026egosteer,zou2026activeglasses,punamiya2026egoverse, chen2026egohumanoid, allshire2025visual, cai2024disentangled, weng2025hdmi,cai2025fastddhpose,lang2025beyond}. They require accurate pose estimation and hand-designed embodiment mappings. \textbf{Implicit} approaches bypass pose recovery and instead learn a shared representation that bridges human and robot domains. Representative strategies include using human video as a large-scale VLM pretraining corpus~\cite{li2025scalable,lin2025physbrain,fu2025metis}, building unified human--robot token spaces through cross-modal reconstruction~\cite{chen2026unit,li2026ace,wang2026mmgait,yin2026mivla}, and deriving latent-action vocabularies from video~\cite{ye2025lapa,zhang2026clap,xu2026motion}. 

However, these methods either share the full action in a single space or entangle it with visual information. The kinematic discrepancy between embodiments and the task-irrelevant noise in raw visual signals both obscure the shared manipulation intent, limiting transferable knowledge extraction. EXE addresses this by factorizing the action into a cross-embodiment positional subspace and an embodiment-specific complement, aligning only the former with an explicit language signal. This avoids pose retargeting and paired demonstrations while preventing domain-specific geometry from contaminating the shared representation.

\section{Method}


EXE transfers manipulation knowledge from egocentric human video to robot control through three stages (Figure~\ref{fig:framework}). \textbf{Stage~1} trains a \emph{factorized action tokenizer} (\S\ref{sec:tokenizer}) that splits each action into a shareable positional subspace and an embodiment-specific kinematic subspace, quantized by separate codebooks, so that motions with the same semantics from different embodiments map to the same shared code. \textbf{Stage~2} performs \emph{cross-embodiment VLM pretraining} (\S\ref{sec:vlm_pretrain}) on both human video and robot demonstrations: the VLM learns to predict the shared action codes from visual and language context. This shapes a cross-embodiment action representation inside the VLM backbone. \textbf{Stage~3} trains a lightweight \emph{action expert} (\S\ref{sec:action_expert}) on downstream robot data with continuous action regression, warm-started from the Stage~2 checkpoint and conditioned on its cross-embodiment representations via \emph{Bridge Features}.

\subsection{Stage1: Training Factorized Action Tokenizer}
\label{sec:tokenizer}

\input{figure/fig_tokenizer}

The tokenizer converts a continuous action chunk into discrete latent codes. Its key design is \emph{feature-group routing}: instead of quantizing the whole action jointly, we split it by cross-embodiment motion similarity, route the transferable part to a shared codebook and keep the rest in per-domain codebooks. Two cross-domain losses further align the shared codebook so that semantically equivalent motions from different embodiments converge onto common code entries.

\paragraph{Action factorization and quantization.}
Let $\mathbf{a}\!\in\!\mathbb{R}^{D}$ denote an action vector in a single-timestep. We partition its dimensions into two \emph{disjoint} groups by their semantic role:
\begin{itemize}
    \item \textbf{Shared group} $\mathbf{a}^{(s)}\!\in\!\mathbb{R}^{D_s}$: the 3D position of the wrist (human) or end-effector (robot), expressed in the egocentric camera frame. For bimanual embodiments this becomes $[\mathbf{p}_L \mid \mathbf{p}_R]\!\in\!\mathbb{R}^{6}$. This positional signal encodes spatial task intent---reaching, trajectory following, target approaching---that is invariant to kinematic structure.
    \item \textbf{Specific group} $\mathbf{a}^{(p)}\!\in\!\mathbb{R}^{D_p}$: the remaining dimensions that are inherently embodiment-dependent. For humans this comprises wrist rotation and finger joint angles. For gripper robots it contains end-effector rotation and gripper state. For dexterous-hand robots such as GR1 it is the full native joint-angle vector.
\end{itemize}
The partition $D_s\!+\!D_p\!=\!D$ is defined by a robot/human \emph{routing table}, a static index map that assigns each action dimension to one of the two groups.

Given an action chunk $\mathbf{A}\!\in\!\mathbb{R}^{T\times D}$ of $T$ timesteps, we extract the dimensions belonging to each group according to the routing table, forming $\mathbf{A}^{(s)}\!\in\!\mathbb{R}^{T\times D_s}$ and $\mathbf{A}^{(p)}\!\in\!\mathbb{R}^{T\times D_p}$. Each group passes through its own encoder--quantizer--decoder pipeline:
\begin{align}
\mathbf{z}^{(s)} &= f_{\text{enc}}^{(s)}(\mathbf{A}^{(s)}) \in \mathbb{R}^{T\times d_q},  \label{eq:shared_enc} \\
\mathbf{z}_q^{(s)}, \; \mathbf{c}^{(s)} &= \operatorname{VQ}_{\text{shared}}(\mathbf{z}^{(s)}), \label{eq:shared_vq} \\
\mathbf{z}^{(p)} &= f_{\text{enc}}^{(p)}(\mathbf{A}^{(p)}) \in \mathbb{R}^{T\times d_q}, \label{eq:spec_enc} \\
\mathbf{z}_q^{(p)}, \; \mathbf{c}^{(p)} &= \operatorname{VQ}_{\text{specific}}(\mathbf{z}^{(p)}), \label{eq:spec_vq}
\end{align}
where $f_{\text{enc}}^{(\cdot)}$ is a per-domain encoder consisting of FFT-based frequency augmentation, dilated 1D convolutions, and a Transformer stack~\cite{dong2026actioncodec}. $\operatorname{VQ}_{\text{shared}}$ is a single codebook of size $K_s$ shared across domains and updated from both human and robot data, implemented with residual vector quantization~\cite{oord2017vqvae}, while $\operatorname{VQ}_{\text{specific}}$ is a per-domain codebook of size $K_p$. Both latents and codebook entries are $\ell_2$-normalized before lookup, so code assignment follows cosine similarity. Unlike standard RVQ which applies all residual levels to the \emph{full} action latent, our RVQ operates within each factorized group, so the shared codebook never sees embodiment-specific dimensions. The decoded outputs $(\hat{\mathbf{A}}^{(s)}, \hat{\mathbf{A}}^{(p)})$ are then placed back into their original positions to reconstruct the full action $\hat{\mathbf{A}} \in \mathbb{R}^{T\times D}$.

Human actions are 3D hand poses extracted from egocentric video via the MANO kinematic model~\cite{romero2017mano}, yielding a 102-dimensional vector per timestep for bimanual settings. Robot actions are end-effector poses, e.g., 20-dimensional vectors covering 3D position, 6D rotation, and gripper state per arm. To ensure the shared-group signals from the two domains are geometrically comparable, both human wrist positions and robot end-effector positions are expressed in the egocentric camera coordinate frame. The chunk length is $T$ for both domains, and the tokenizer produces $T$ discrete tokens per chunk with per-step tokenization.

\paragraph{Cross-domain alignment.}
Beyond reconstruction, two complementary losses shape the shared codebook to support cross-embodiment transfer. First, a symmetric InfoNCE loss $\mathcal{L}_{\text{NCE}}$ pulls together the latents shared by human and robot that are quantized to the same shared code while pushing apart latents assigned to different codes. Concretely, let $\mathbf{z}_i^{h}$ and $\mathbf{z}_j^{r}$ denote $\ell_2$-normalized shared-group latents from the human and robot domains, and let $\mathbf{c}_i^{h}$, $\mathbf{c}_j^{r}$ be their assigned shared-code indices. The positive set for anchor $i$ is $\mathcal{P}_i = \{j \mid \mathbf{c}_i^{h} = \mathbf{c}_j^{r}\}$. The loss is:
\begin{equation}
\mathcal{L}_{\text{NCE}} = -\frac{1}{2}\left( \mathbb{E}_{i}\left[\frac{1}{|\mathcal{P}_i|}\sum_{j\in\mathcal{P}_i} \log \frac{e^{\cos(\mathbf{z}_i^{h},\mathbf{z}_j^{r})/\tau}}{\sum_{k} e^{\cos(\mathbf{z}_i^{h},\mathbf{z}_k^{r})/\tau}}\right] + \text{sym} \right),
\label{eq:nce}
\end{equation}
where $\tau$ is a temperature hyperparameter and ``sym'' denotes the symmetric term with robot as anchor. This encourages the shared codebook to develop consistent code semantics across domains.

However, contrastive alignment alone faces a bootstrapping problem. Since code assignment is determined by spatial trajectory, human and robot actions for the same task are typically routed to \emph{different} codes when their trajectories differ spatially, preventing them from ever becoming contrastive positives. Language breaks this problem by providing an external semantic signal invariant to embodiment geometry. We use a frozen Qwen3-Embedding-8B~\cite{zhang2025qwen3} model $\phi$ to encode each sample's task instruction into a text embedding $\mathbf{e} = \phi(L)$, precomputed offline. During each dual-domain step, we form two cross-domain similarity matrices with human samples indexed by $i$ and robot samples by $j$:
\begin{equation}
S_{\text{text}}[i,j] = \cos\bigl(\mathbf{e}_i^{h},\; \mathbf{e}_j^{r}\bigr), \quad
S_{\text{act}}[i,j] = \cos\bigl(\bar{\mathbf{z}}_i^{h},\; \bar{\mathbf{z}}_j^{r}\bigr),
\label{eq:sim_matrices}
\end{equation}
where $\bar{\mathbf{z}} = \frac{1}{T}\sum_{t}\mathbf{z}_t^{(s)}$ is the temporal mean of the per-chunk shared latent. The text matrix $S_{\text{text}}$ is detached and serves as a soft target. The relational distillation loss is:
\begin{equation}
\mathcal{L}_{\text{lang}} = \bigl\lVert S_{\text{act}} - S_{\text{text}} \bigr\rVert_F^2.
\label{eq:lang_loss}
\end{equation}
This loss is computed only over cross-domain sample pairs to avoid the degenerate within-domain case where $S_{\text{text}} \equiv 1$. Language never enters the encoder input---it acts solely as a distributional shaping signal for the shared latent space and is discarded after tokenizer training.

\input{figure/fig_framework}

\paragraph{Training objective.}
The tokenizer is jointly trained on human and robot data using dual-domain steps. Each step draws a batch from each domain, runs both through the shared and specific pipelines, and updates all parameters in a single backward pass. The per-domain loss decomposes into shared-group and specific-group terms:
\begin{align}
\mathcal{L}_{\text{tok}} =\;& \mathcal{L}_{\text{recon}}^{(s)} + \alpha_s \mathcal{L}_{\text{VQ}}^{(s)} + \gamma_s \mathcal{L}_{\text{div}}^{(s)} \notag \\
& + \mathcal{L}_{\text{recon}}^{(p)} + \alpha_p \mathcal{L}_{\text{VQ}}^{(p)} + \beta \mathcal{L}_{\text{mask}}^{(p)} + \gamma_p \mathcal{L}_{\text{div}}^{(p)},
\label{eq:tok_loss}
\end{align}
where $\mathcal{L}_{\text{recon}}^{(s)}$ and $\mathcal{L}_{\text{recon}}^{(p)}$ are MSE reconstruction losses on the shared and specific subspaces respectively, $\mathcal{L}_{\text{VQ}}^{(\cdot)}$ denotes the VQ commitment loss~\cite{van2017neural} for each codebook, $\mathcal{L}_{\text{mask}}^{(p)}$ is a masked-latent consistency loss on the specific-group latent that enhances temporal robustness of the encoder~\cite{dong2026actioncodec}, and $\mathcal{L}_{\text{div}}^{(\cdot)}$ is a soft-entropy diversity regularizer encouraging codebook utilization. The complete Stage~1 objective combines per-domain reconstruction with cross-domain alignment:
\begin{equation}
\mathcal{L}_1 = \mathcal{L}_{\text{tok}}^{h} + \mathcal{L}_{\text{tok}}^{r} + \lambda_c \mathcal{L}_{\text{NCE}} + \lambda_l \mathcal{L}_{\text{lang}}.
\label{eq:stage1_loss}
\end{equation}
After Stage~1, we obtain one shared codebook $\mathcal{C}_s$ of size $K_s$ capturing the cross-embodiment spatial-intent vocabulary and one specific codebook $\mathcal{C}_p^d$ per domain $d$. All tokenizer weights are frozen for subsequent stages.

\subsection{Stage2: Cross-Embodiment VLM Pretraining}
\label{sec:vlm_pretrain}

In Stage~2, our goal is to transfer the cross-embodiment action knowledge captured by the shared codebook into the VLM backbone. We train the VLM to predict the discrete action codes learned in Stage~1, where the model maps an image and a language instruction to shared-codebook indices.

\paragraph{Architecture}
The VLM backbone is Florence-2-large~\cite{xiao2023florence}, whose LM head vocabulary is extended with $K_s$ entries corresponding to the shared codebook. We append $T$ special placeholder tokens to the text prompt, and the LM head is trained to predict the shared-code index at each placeholder position.

\paragraph{Joint dual-domain training}
Stage~2 interleaves batches from two data sources: egocentric human video with paired hand-pose actions~\cite{li2026egolive,li2025scalable,hoque2026egodex} and robot manipulation demonstrations. Training steps alternate between the two domains at a configurable ratio. At each step, the frozen Stage~1 tokenizer encodes the action chunk into discrete code indices as ground truth targets. For human data, the VLM is trained to predict only the shared codes $\mathbf{c}^{(s)}$. For robot data, the VLM predicts both the shared codes and the robot-specific codes $\mathbf{c}^{(p)}$, since our ultimate goal is to let the robot acquire human-like motion patterns through the shared vocabulary while retaining its own embodiment-specific control. The training loss is:
\begin{equation}
\mathcal{L}_2 = \mathcal{L}_{\text{CE}}^{(s)} + \mathbf{1}_{\text{robot}} \cdot \mathcal{L}_{\text{CE}}^{(p)},
\label{eq:stage2_loss}
\end{equation}
where $\mathbf{1}_{\text{robot}}$ indicates robot-domain steps. The action expert is \emph{not used} during this stage, so Stage~2 focuses exclusively on shaping the VLM's internal representations.


\subsection{Stage3: VLA Fine-tuning}
\label{sec:action_expert}

Stage~3 converts the pretrained VLM representations into executable continuous robot actions through a dedicated action expert, using only downstream robot data.

\paragraph{Training.}
We load the Stage~2 VLM checkpoint and attach an action transformer as the action expert. At each step, the VLM processes the image and instruction to produce a sequence of hidden states, which serve as conditioning context. To explicitly reuse the cross-embodiment representations learned in Stage~2, we further extract the VLM hidden states at the action-placeholder positions as \emph{Bridge Features}, and concatenate them with the VLM features fed to the action expert. The action transformer takes noisy action tokens as input and attends over these features to decode continuous robot actions $\hat{\mathbf{A}} \in \mathbb{R}^{T\times D}$. The discrete tokenizer and all latent-action code supervision are removed entirely. The training loss is purely a continuous action regression $\mathcal{L}_3 = \mathcal{L}_{\text{action}}$, where $\mathcal{L}_{\text{action}}$ is the action regression objective following~\cite{zheng2026x}. The cross-embodiment knowledge acquired during Stage~2 persists in the VLM backbone's weights, and the action expert learns to decode these representations into robot-specific actions through end-to-end fine-tuning.

\paragraph{Inference.}
Given an image and a language instruction, the VLM produces the conditioning hidden states, and the action transformer decodes continuous robot actions from sampled noise. No tokenizer, codebook, or language model is involved: the discrete action vocabulary is purely a training-time mechanism in Stages~1--2 and is absent at deployment.

\input{figure/fig_tsne}
\section{Experiments}


\subsection{Implementation Details}
\label{sec:exp_impl}

In Stage~1, we train the factorized tokenizer for $50$k steps with a shared codebook of size $64$, a per-domain specific codebook of size $64$, and a quantization dimension of $32$. Stages~2 and~3 are built on the X-VLA~\cite{zheng2026x} framework and use the Florence-2-large~\cite{xiao2023florence} backbone. In Stage~2, we pretrain the backbone with a cross-entropy loss to predict shared action tokens from visual and language inputs for $30$k iterations, deliberately excluding large-scale cross-embodiment robot data to isolate the effect of egocentric human video pretraining. In Stage~3, we randomly initialize the action expert and post-train it on robot data for $60$k iterations, inheriting the representations learned in Stage~2. All stages use a batch size of $32$ with a learning rate of $6\mathrm{e}{-4}$ and $20$k warmup steps, and we train through behavior cloning with an $\ell_1$ loss on eight NVIDIA A800 GPUs.

\subsection{Benchmark and Metrics} 
\label{sec:exp_robotwin}

\paragraph{RoboTwin 2.0}
We evaluate on RoboTwin 2.0~\cite{chen2025robotwin2}, a bimanual manipulation benchmark based on an ALOHA~\cite{zhao2023learning} dual-arm robot with 50 diverse tabletop tasks. Each task is tested under an \emph{easy} split with fixed scenes and a \emph{hard} split with randomized backgrounds, lighting, and object layouts. We run 50 episodes per task per split and report the average success rate across both splits.

\paragraph{RoboCasa GR1}
We also evaluate on dexterous-hand humanoid manipulation benchmark RoboCasa GR1~\cite{nvidia2025gr00tn1openfoundation}. Its 24 tasks cover household interactions with articulated objects and multi-stage goals, presenting greater embodiment complexity than typical single-arm benchmarks. Each task provides roughly 1{,}000 demonstrations, yielding about 24K trajectories in total.
\paragraph{Real-world evaluation}
To further validate that egocentric human video pretraining transfers beyond simulation, we conduct real-world experiments on an AgileX Piper bimanual platform. We evaluate the ``sweep the rubbish'' task, for which we collect 2{,}450 teleoperated real-world demonstrations as the downstream robot dataset. The Stage~3 action expert is trained on this data for 30k steps with a batch size of 32, warm-started from the Stage~2 VLM checkpoint. At deployment, the policy is rolled out over 10 trials on the physical platform, and we report the success rate.

\paragraph{Baseline}
To isolate the effect of our pretraining, we construct an ablated variant of EXE, denoted \textbf{EXE\textsubscript{base}}, which uses the same X-VLA~\cite{zheng2026x} framework and Florence-2-large~\cite{xiao2023florence} backbone, but trains the Stage~3 action expert directly on downstream robot data. It does not load the pretrained weights from the large-scale robot teleoperation pretraining of the original X-VLA, and omits our Stage~1 and Stage~2 pretraining. The improvement of EXE over EXE\textsubscript{base} therefore directly reflects the benefit of these two stages.

\subsection{Tokenizer Analysis}
\label{sec:exp_tokenizer}

In this section, we examine whether the factorized tokenizer produces the intended structure in both the shared and embodiment-specific latent spaces: (1)~cross-domain alignment in the shared space and clear domain separation in the embodiment-specific space, and (2)~diverse, non-collapsed codebook utilization.

\paragraph{Latent space visualization}
To verify whether the shared codebook pulls the two domains onto the same manifold at the feature level, we pass real actions from the human~\cite{li2026egolive} and robot~\cite{chen2025robotwin2,nvidia2025gr00tn1openfoundation} domains through the trained Stage~1 tokenizer, quantize them with the shared-group codebook, and use the $\ell_2$-normalized codebook vectors $\bar{\mathbf{z}}^{(s)}_{q}$ as features. We randomly sample features from each domain and perform joint t-SNE with cosine distance (Fig.~\ref{fig:tsne}). Blue denotes human data and red denotes robot data; colors also indicate discrete code ids.

In Fig.~\ref{fig:tsne}, shared-group tokens from both domains overlap in compact clusters. Tokens assigned to the same code id form local groups, showing that both embodiments genuinely share a common action vocabulary. In contrast, the specific-group tokens shows clear domain separation: the per-domain specific codebooks are trained independently, so their tokens form two distinct clusters. This matches the intended design, where the specific group encodes domain-private semantics (rotation, gripper, hand joints) that need not be shared.

We also visualize the 3D end-effector/wrist trajectories projections behind shared codes in Fig.~\ref{fig:tsne}. We select a shared action code that is activated by both the human and the robot within a single episode, and plot the corresponding end-effector trajectory segments. The two segments trace out visibly similar motions: both correspond to a top-to-bottom placing action that lowers the end-effector to set down the grasped object. This confirms that a single shared code indeed grounds to the same physical spatial intent in both embodiments, providing a concrete, trajectory-level counterpart to the cluster-level sharing observed in Fig.~\ref{fig:tsne}.

\input{table/codebook_usage}

\paragraph{Shared codebook utilization}
We analyze Stage~1 codebook usage in Table~\ref{tab:codebook_usage}. Both codebooks activate nearly all codes, and the co-activated codes in the shared codebook are also almost entirely used by both human and robot domains, indicating that actions from different embodiments indeed share common motion patterns in the latent space.



\input{table/data_source}

\subsection{Main Results}
\label{sec:exp_main_results}
A central question for our approach is whether egocentric human video pretraining through EXE's cross-embodiment pipeline can improve downstream robot performance. We use EgoLive as the human data source for Stage~1 tokenizer training and Stage~2 VLM co-training, and evaluate the transferred policy on the three benchmarks introduced in \S\ref{sec:exp_robotwin}: RoboTwin~2.0, RoboCasa~GR1, and real-world tasks.

\subsubsection{Results on RoboTwin 2.0}
\label{sec:exp_robotwin_results}

We use RoboTwin 2.0 robot data as the robot domain and fine-tune Stage~3 on RoboTwin. Table~\ref{tab:data_source} compares EXE against prior VLA baselines. EgoLive pretraining yields consistent improvements over EXE\textsubscript{base}, and EXE reaches an average success rate of 76.2\% (75.6\% easy, 76.8\% hard), outperforming SmolVLA~\cite{shukor2025smolvla}, X-VLA~\cite{zheng2026x}, and TurboVLA~\cite{xie2026turbovla}. This confirms that EXE's factorized tokenizer and co-training pipeline can extract transferable manipulation knowledge from egocentric human video and translate it into gains on bimanual tabletop manipulation.

\subsubsection{Results on RoboCasa GR1}
\label{sec:exp_robocasa_results}

We use the StarVLA~\cite{community2026starvla} framework to train GR1 on the 24 RoboCasa tasks for 100k steps in Stage 3. Table~\ref{tab:robocasa_gr1} reports the average success rate. EXE reaches 56.6\%, improving over both EXE\textsubscript{base} (52.0\%) and the official result of StarVLA-GR00T (47.8\%). This indicates that the shared spatial-intent vocabulary learned from human video not only benefits gripper-based manipulation but also improves success rates on dexterous hand tasks.

\subsubsection{Results in the Real World}
\label{sec:exp_realworld_results}

We validate transferability beyond simulation on the AgileX Piper bimanual platform (Figure~\ref{fig:real_world}). The Stage~3 action expert is warm-started from the Stage~2 VLM checkpoint and fine-tuned on 2{,}450 teleoperated ``sweep the rubbish'' demonstrations. We design two evaluation groups, each consisting of 10 rollouts for both EXE\textsubscript{base} and EXE model:

\begin{itemize}
    \item \textbf{In-distribution}: objects and positions seen during training. EXE\textsubscript{base} achieves \textbf{60\%} and EXE achieves \textbf{80\%}. The gain is most pronounced on visually colorful objects---e.g., a broom-and-dustpan set with distinctly colored parts---suggesting that egocentric pretraining helps the policy leverage rich color cues for more accurate manipulation.
    \item \textbf{Putting unseen objects}: novel objects not present in the training set are placed in the scene as distractors. EXE\textsubscript{base} achieves \textbf{10\%} while EXE achieves \textbf{30\%}, indicating stronger scene-level generalization under visual distractions introduced by unseen objects.
\end{itemize}

\input{table/combined_robocasa_cotrain}

\input{figure/fig_real_world}

\subsection{Ablation Study}
\label{sec:exp_ablation}

\input{table/ablation}

\paragraph{Component ablation.}
We isolate the contribution of each EXE component and report results in Table~\ref{tab:ablation}. The first row is the EXE\textsubscript{base} baseline, yielding 73.2\%.

The second and third rows ablate the alignment losses in Stage~1 tokenizer training. Adding the shared-code InfoNCE loss raises the average to 73.8\%. This loss pulls human and robot latents assigned to the same code closer, stabilizing the shared codebook. Further adding language-guided relational distillation improves the average to 74.1\%. Because the frozen language embedding is embodiment-invariant, it guides semantically equivalent actions onto nearby codes even when their spatial trajectories differ---the regime where the purely spatial InfoNCE has no positive pairs.

The last row adds the Bridge Feature, which verifies whether the shared action token features obtained from Stage~2 co-training benefit Stage~3 fine-tuning. Enabling it brings the full EXE pipeline to 76.2\%, confirming that the cross-embodiment knowledge encoded in the shared tokens transfers effectively to downstream policy learning. Each component contributes independently, and removing any one degrades performance.

\paragraph{Co-training Data Comparison}
We study how the mixing ratio between egocentric human~\cite{li2026egolive} and robot~\cite{chen2025robotwin2} data in Stage~2 co-training affects downstream performance, and report the results in Table~\ref{tab:cotrain_ratio}. A balanced 1:1 ratio achieves the best average success rate (76.2\%), outperforming the robot-heavy 2:1 (74.3\%) and 3:1 (75.5\%) mixtures. This suggests that maintaining sufficient human data exposure during co-training is important for preserving the cross-embodiment knowledge acquired in Stage~1, whereas over-weighting robot data biases the backbone toward the robot domain and erodes the transferable shared representation.

\paragraph{Generalization across human data sources.}
To verify that EXE is not tied to a specific egocentric dataset, we fix the robot domain (RoboTwin~2.0) and vary the human data source used in Stage~1 tokenizer training and Stage~2 VLM co-training. Table~\ref{tab:human_dataset} reports the results with three representative datasets: VITRA-1M~\cite{li2025scalable}, EgoDex~\cite{hoque2026egodex}, and EgoLive~\cite{li2026egolive}. All three sources yield improvements over EXE\textsubscript{base}, confirming that EXE's factorized tokenizer and cross-embodiment pipeline can extract transferable manipulation knowledge from diverse egocentric video sources. Among them, EgoLive achieves the best average success rate (76.2\%), likely because its naturalistic hand-manipulation scenarios provide richer spatial-intent coverage that aligns well with the bimanual RoboTwin 2.0 benchmark. These results demonstrate that EXE generalizes robustly across different human data distributions.

\input{table/human_dataset}

\section*{Conclusion}

We presented EXE, an egocentric cross-embodiment pretraining framework that factorizes actions into a shared positional subspace and an embodiment-specific kinematic subspace, enabling manipulation semantics transfer from human videos to robots with grippers or dexterous hands, offering a general recipe for cross-embodiment pretraining regardless of the robot type. A dual alignment combining contrastive learning with language-guided relational distillation ensures that cross-embodiment actions sharing both spatial similarity and task semantics converge in the shared space. Experiments on simulated and real-world tasks demonstrate consistent improvements, validating that factorized action representations effectively transfer manipulation knowledge from egocentric human video to robot policies.



\bibliographystyle{ieeetr}
\bibliography{root}

\end{document}

%% file: figure/fig_tokenizer.tex
\begin{figure}[t]
  \centering
  \includegraphics[width=\columnwidth]{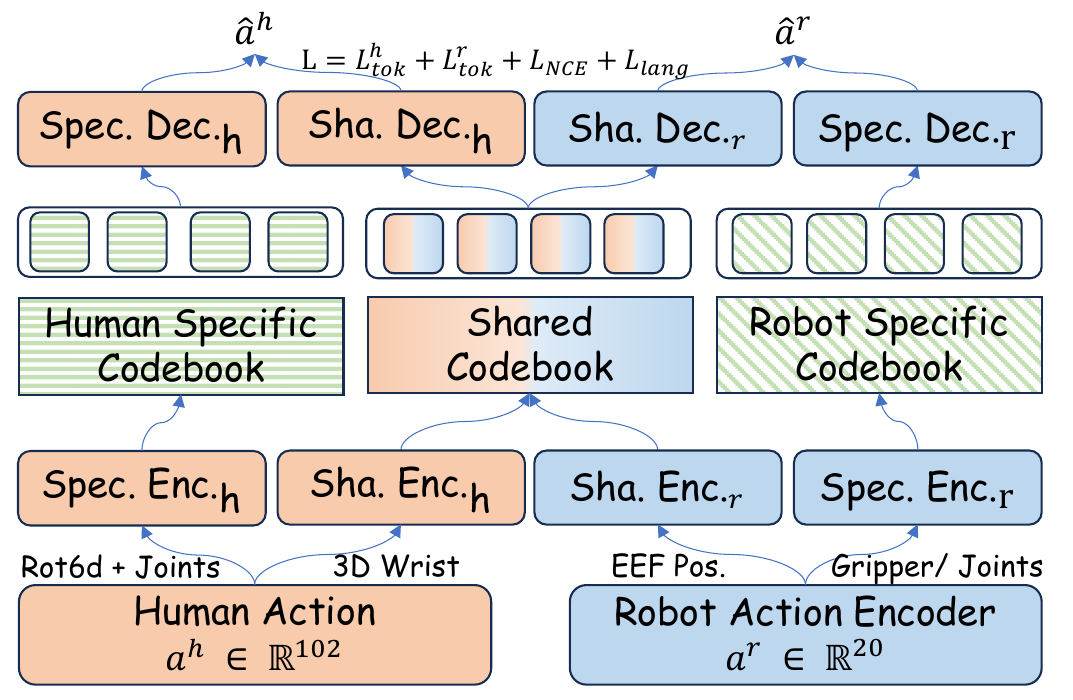}
  \caption{%
    \textbf{Factorized Action Tokenizer.}
    Sha./Spec. Enc.\ and Dec.\ denote the shared/specific encoder and
    decoder. The shared group uses one cross-domain
    codebook, the specific group per-domain codebooks.
  }
  \vspace{-4mm}
  \label{fig:tokenizer}
\end{figure}

%% file: figure/fig_framework.tex
\begin{figure}[t]
  \centering
  \includegraphics[width=\columnwidth]{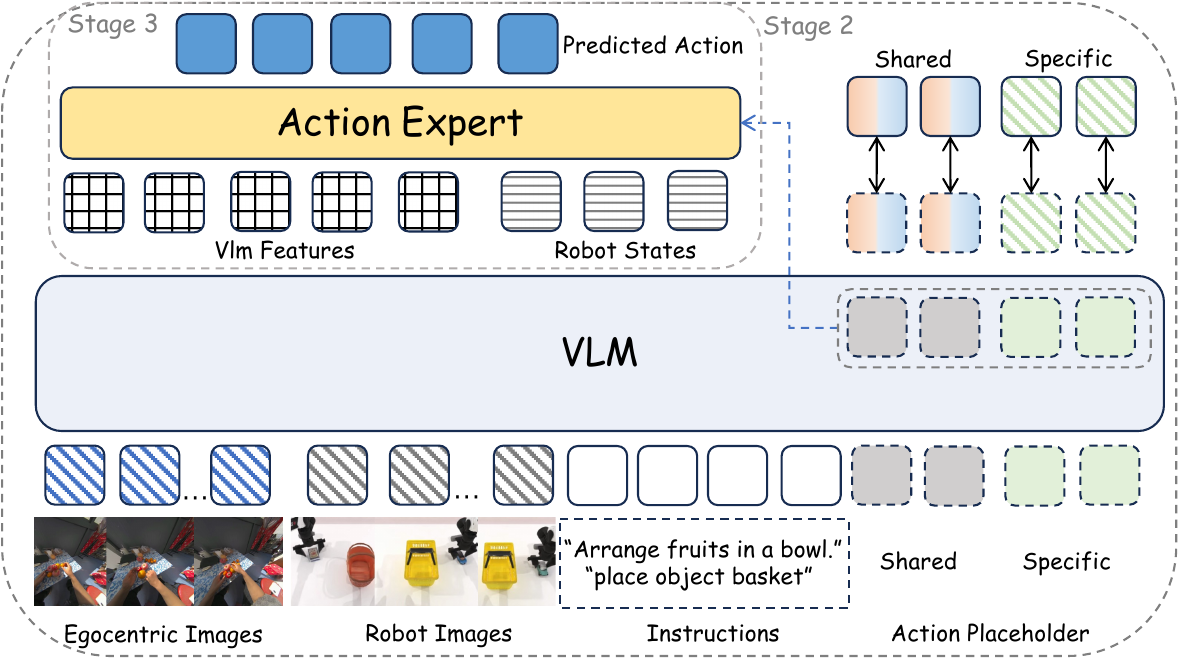}
  \caption{%
    \textbf{Overview of EXE.}
    The VLM takes egocentric/robot images, instructions, and
    shared/specific action placeholders (dashed boxes) as input.
    \textbf{Stage~2} predicts action codes at the placeholder positions;
    \textbf{Stage~3} decodes VLM features into continuous actions.
  }
  \label{fig:framework}
\end{figure}

%% file: figure/fig_tsne.tex
\begin{figure*}[t]
  \centering
  \includegraphics[width=\textwidth]{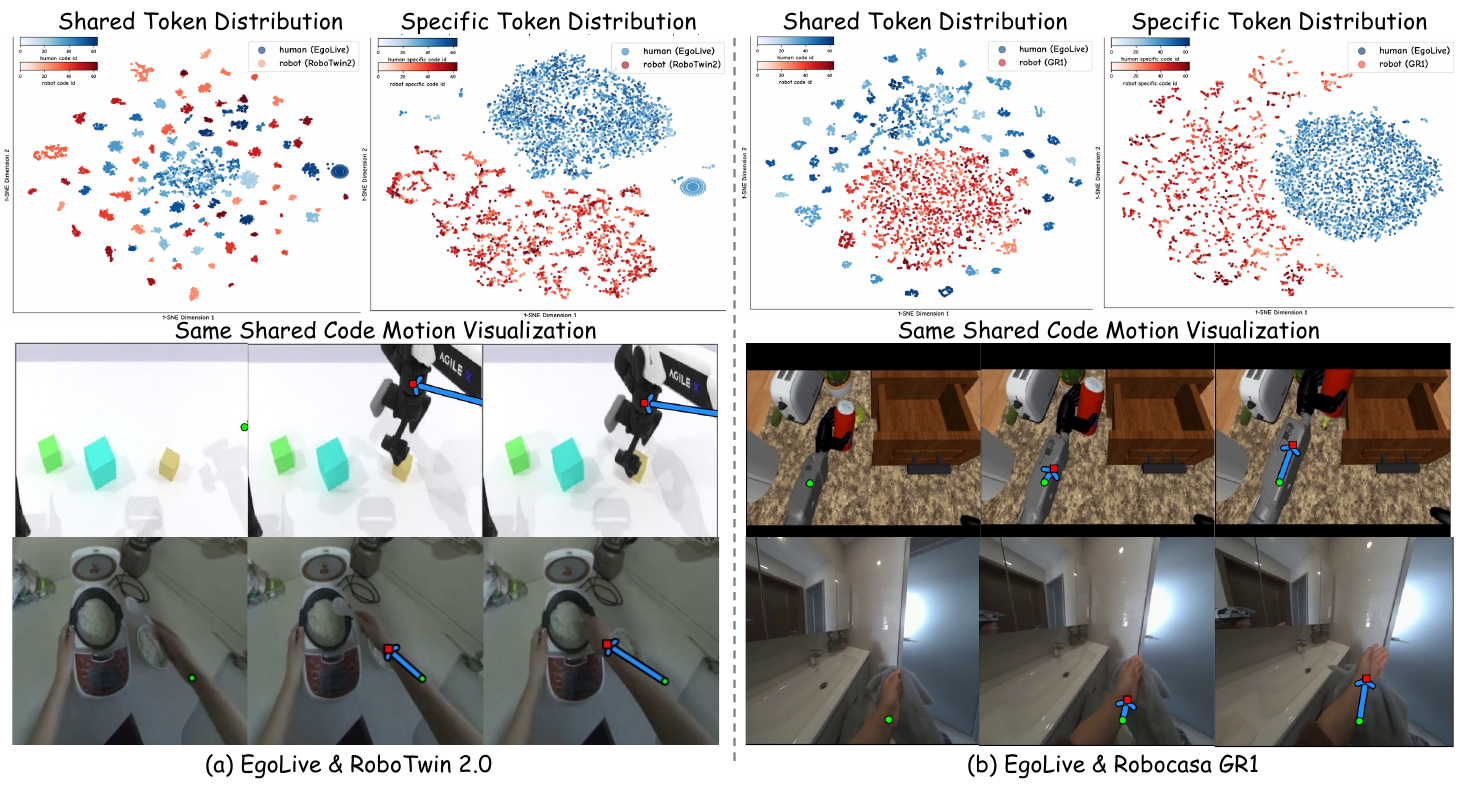}
  \vspace{-8mm}
  \caption{
    \textbf{(a)}~EgoLive \& RoboTwin~2.0 and \textbf{(b)}~EgoLive \& Robocasa GR1.
    Each panel shows three views: the \emph{Shared Token Distribution}, the \emph{Specific Token Distribution}, and the absolute 3D end-effector/wrist trajectories projections
    of a robot chunk (\emph{top}) and a human chunk (\emph{bottom}) quantized to the same shared code, where both moves to the same direction (green = start, red = end).
  }
  \label{fig:tsne}
\end{figure*}

%% file: table/codebook_usage.tex
\begin{table}[htbp]
  \centering
  \caption{%
    Stage~1 codebook utilization. ``Co'' means the number of codes co-actived in both domains. H refers to Human domain, R refers to RobotWin2.0, G refers to Robocasa GR1.
  }
  \label{tab:codebook_usage}
  \setlength{\tabcolsep}{4pt}
  \fontsize{9pt}{11pt}\selectfont
  \begin{tabular}{c ccc| ccc}
    \toprule
    Type & Domain & Actived & Co & Domain & Actived & Co \\
    \midrule
    Shared
             & H \& R & 59 & 58 & H \& R & 64 & 64 \\
    \multirow{2}{*}{Specific}
             & H & 64 & -- & H & 64 & -- \\
             & R & 64 & -- & G & 64 & -- \\
    \bottomrule
  \end{tabular}
\end{table}

%% file: table/data_source.tex
\begin{table}[t]
  \centering
  \caption{%
    Evaluation Results (\%) on RoboTwin 2.0. RPT: robot-pretraining on large-scale teleoperated data. EXE\textsubscript{base} is an ablated EXE without our Stage~1--2 pretraining.
  }
  \label{tab:data_source}
  \setlength{\tabcolsep}{4pt}
  \fontsize{9pt}{11pt}\selectfont
  \begin{tabular}{lccccc}
    \toprule
    Model & Params & RPT & Easy & Hard & AVG \\
    \midrule
    SmolVLA~\cite{shukor2025smolvla} & 0.5B & $\times$ & 59.2 & 60.8 & 60.0 \\
    X-VLA~\cite{zheng2026x} & 0.8B & $\surd$ & 72.9 & 72.8 & 72.9 \\
    TurboVLA~\cite{xie2026turbovla} & 0.4B & $\times$ & -- & -- & 60.2 \\
    EXE\textsubscript{base} & 0.8B & $\times$ & 68.4 & 65.2 & 66.8 \\
    EXE & 0.8B & $\times$ & \textbf{75.6} & \textbf{76.8} & \textbf{76.2} \\
    \bottomrule
  \end{tabular}
\end{table}

%% file: table/combined_robocasa_cotrain.tex
\begin{table}[t]
  \noindent
  \begin{minipage}[t]{0.47\columnwidth}
    \raggedright
    \caption{%
      Evaluation Results (\%) on RoboCasa GR1.
    }
    \label{tab:robocasa_gr1}
  \end{minipage}%
  \hspace{0.02\columnwidth}%
  \begin{minipage}[t]{0.47\columnwidth}
    \raggedright
    \caption{%
      Stage~2 co-training ratio of human to robot data (\%).
    }
    \label{tab:cotrain_ratio}
  \end{minipage}
  \par\noindent
  \setlength{\tabcolsep}{4pt}
  \fontsize{9pt}{11pt}\selectfont
  \begin{tabular}{@{}lc@{}c@{}lccc@{}}
    \cmidrule[\heavyrulewidth]{1-2}\cmidrule[\heavyrulewidth]{4-7}
    Method & AVG & \hspace{0.04\columnwidth} & Ego:Robot & Easy & Hard & AVG \\
    \cmidrule[\lightrulewidth]{1-2}\cmidrule[\lightrulewidth]{4-7}
    StarVLA-GR00T~\cite{community2026starvla} & 47.8 & & 1:1 & \textbf{75.6} & \textbf{76.8} & \textbf{76.2} \\
    EXE\textsubscript{base} & 52.0 & & 2:1 & 75.5 & 73.2 & 74.3 \\
    EXE & \textbf{56.6} & & 3:1 & 76.0 & 75.0 & 75.5 \\
    \cmidrule[\heavyrulewidth]{1-2}\cmidrule[\heavyrulewidth]{4-7}
  \end{tabular}
\end{table}

%% file: figure/fig_real_world.tex
\begin{figure}[t]
  \centering
  \includegraphics[width=\columnwidth]{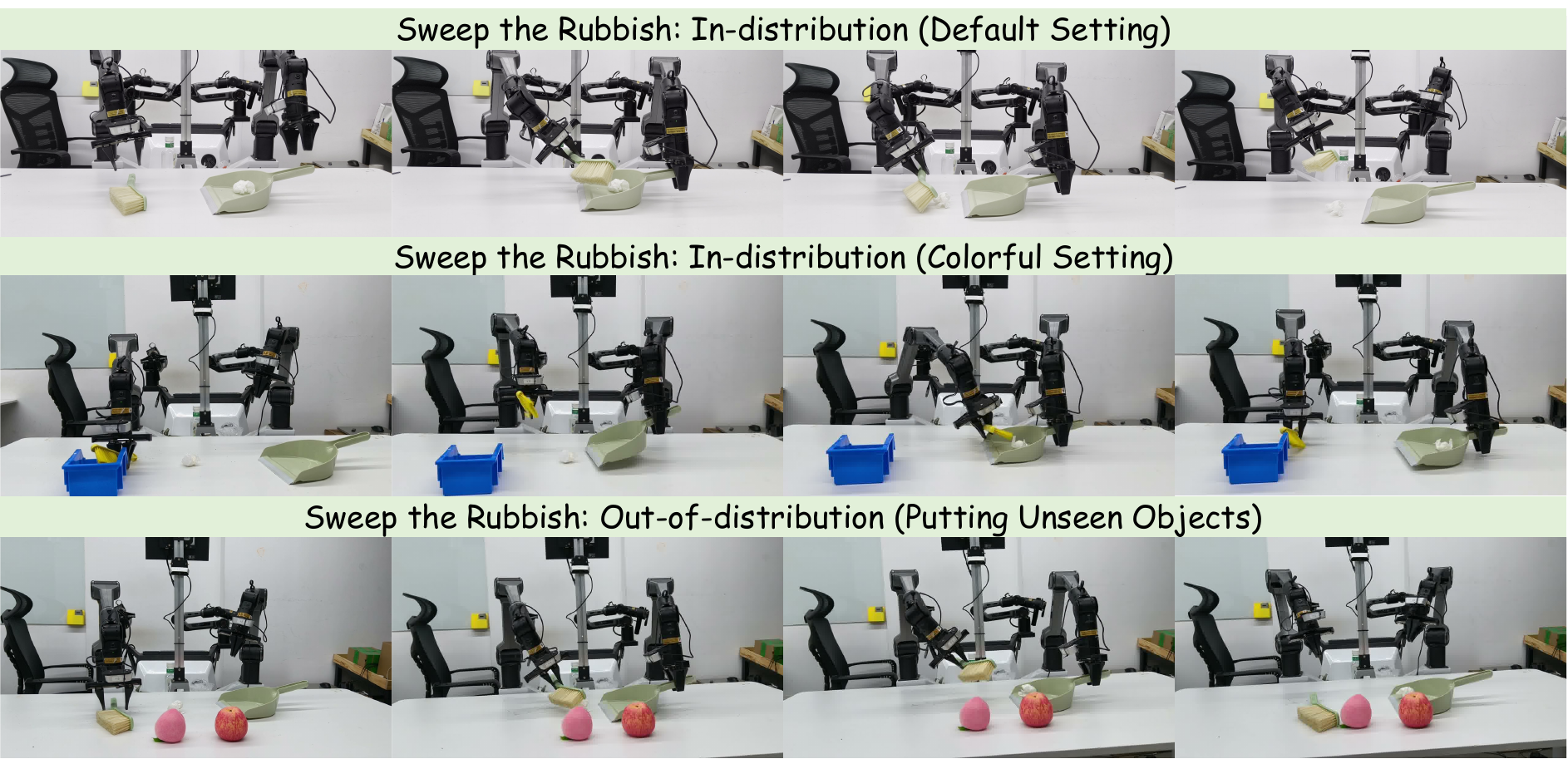}
  \caption{%
    Real-world evaluation on the AgileX Piper bimanual platform.
  }
  \label{fig:real_world}
\end{figure}

%% file: table/ablation.tex
\begin{table}[t]
  \centering
  \caption{%
    Ablation of EXE components on RoboTwin success rate (\%).
  }
  \label{tab:ablation}
  \setlength{\tabcolsep}{4pt}
  \fontsize{9pt}{11pt}\selectfont
  \begin{tabular}{cccccc}
    \toprule
    \multicolumn{2}{c}{Stage~1} & Stage~3 & \multirow{2}{*}{Easy} & \multirow{2}{*}{Hard} & \multirow{2}{*}{AVG} \\
    \cmidrule(lr){1-2} \cmidrule(lr){3-3}
    InfoNCE & Language Align & Bridge Feature & & & \\
    \midrule
    $\times$ & $\times$ & $\times$ & 74.1 & 72.3 & 73.2 \\
    $\checkmark$ & $\times$ & $\times$ & 74.6 & 73.0 & 73.8 \\
    $\checkmark$ & $\checkmark$ & $\times$ & 75.2 & 73.0 & 74.1 \\
    $\checkmark$ & $\checkmark$ & $\checkmark$ & \textbf{75.6} & \textbf{76.8} & \textbf{76.2} \\
    \bottomrule
  \end{tabular}
\end{table}

%% file: table/human_dataset.tex
\begin{table}[t]
  \centering
  \caption{%
    Generalization across human datasets (\%).
  }
  \label{tab:human_dataset}
  \setlength{\tabcolsep}{4pt}
  \fontsize{9pt}{11pt}\selectfont
  \begin{tabular}{lcccc}
    \toprule
    Human Dataset & Robot Dataset & Easy & Hard & AVG \\
    \midrule
    VITRA-1M~\cite{li2025scalable} & RoboTwin~2.0 & 75.2 & 75.2 & 75.2 \\
    EgoDex~\cite{hoque2026egodex} & RoboTwin~2.0 & 75.2 & 76.4 & 75.8 \\
    EgoLive~\cite{li2026egolive} & RoboTwin~2.0 & \textbf{75.6} & \textbf{76.8} & \textbf{76.2} \\
    \bottomrule
  \end{tabular}
\end{table}